\RequirePackage{fix-cm}

\documentclass[twocolumn]{svjour3}          % twocolumn

\smartqed  % flush right qed marks, e.g. at end of proof

\usepackage[misc]{ifsym}

\usepackage[colorlinks,linkcolor=red,anchorcolor=green,citecolor=blue]{hyperref}

\usepackage{epstopdf}
\usepackage{makecell}
\usepackage{graphicx}	% 插入图片searchtime
\usepackage{float}
\usepackage{color}
\usepackage{arydshln}
\usepackage[section]{placeins}
\usepackage{caption}
\usepackage{tabularx}
\usepackage{booktabs}
\usepackage{amsmath,amsfonts,amssymb,bm}
\usepackage{multirow}
\usepackage{diagbox}
\usepackage{makecell}
\usepackage[ruled]{algorithm2e}		% 插入算法框架
\usepackage{subfigure} 				% 插入子图
\usepackage{autobreak}

\usepackage[authoryear]{natbib}
\begin{document}

\title{A Multi-objective Memetic Algorithm for Auto Adversarial Attack Optimization Design} 

%\author{}
\author{Jialiang Sun \and
        Wen Yao \textsuperscript{\space \Letter} \and
        Tingsong Jiang \textsuperscript{\space \Letter} \and
        Xiaoqian Chen \thanks{Corresponding author: Wen Yao \and Tingsong Jiang
        }
    }
%}
%\author{Xiaoqian Chen\textsuperscript{1} \and
%	Xianqi Chen\textsuperscript{2} \and
%	Weien Zhou\textsuperscript{1} \and
%	Jun Zhang\textsuperscript{1} \and
%	Wen Yao\textsuperscript{1,*}
%}

%\authorrunning{J. Sun et al.} % if too long for running head
   
%\institute{
%	\Letter \quad Wen Yao \at
%	\email{wendy0782@126.com} \\
%	\and
%	Xiaoqian Chen \at
%	\email{chenxiaoqian@nudt.edu.cn} \\
%	\and
%	Xianqi Chen \at
%	\email{chenxianqi12@nudt.edu.cn} \\	
%	\at 
%	\textsuperscript{1} National Innovation Institute of Defense Technology, Chinese Academy of Military Science, Beijing 100000, China \\ \and
%	\textsuperscript{2} College of Aerospace Science and Engineering, National University of Defense Technology, Changsha 410073, China \\
%}

\institute{Jialiang Sun \at
        Defense Innovation Institute, Chinese Academy of Military Science, Beijing 100000, China \\
        \email{sun1903676706@163.com}           %  \\
           \and
         \Letter \quad  Wen Yao \at
              Defense Innovation Institute, Chinese Academy of Military Science, Beijing 100000, China \\ 
             \email{wendy0782@126.com}
           \and
	        \Letter \quad   Tingsong Jiang \at
	Defense Innovation Institute, Chinese Academy of Military Science, Beijing 100000, China \\
	\email{tingsong@pku.edu.cn}
	           \and
	Xiaoqian Chen \at
	Defense Innovation Institute, Chinese Academy of Military Science, Beijing 100000, China \\
	\email{chenxiaoqian@nudt.edu.cn}
}

\date{Received: date / Accepted: date}

\maketitle

\begin{abstract}
	The phenomenon of adversarial examples has been revealed in variant scenarios. Recent studies show that well-designed adversarial defense strategies can improve the robustness of deep learning models against adversarial examples. However, with the rapid development of defense technologies, it also tends to be more difficult to evaluate the robustness of the defensed model due to the weak performance of existing manually designed adversarial attacks. To address the challenge, given the defensed model, the efficient adversarial attack with less computational burden and lower robust accuracy is needed to be further exploited. Therefore, we propose a multi-objective memetic algorithm for auto adversarial attack optimization design, which realizes the automatical search for the near-optimal adversarial attack towards defensed models. Firstly, the more general mathematical model of auto adversarial attack optimization design is constructed, where the search space includes not only the attacker operations, magnitude, iteration number, and loss functions but also the connection ways of multiple adversarial attacks. In addition, we develop a multi-objective memetic algorithm combining NSGA-II and local search to solve the optimization problem. Finally, to decrease the evaluation cost during the search, we propose a representative data selection strategy based on the sorting of cross entropy loss values of each images output by models. Experiments on CIFAR10, CIFAR100, and ImageNet datasets show the effectiveness of our proposed method.

\end{abstract}

\section{Introduction}

With the rapid development of deep learning techniques, deep neural networks (DNN) have been applied widely in a majority of tasks such as image classification \cite{Remote2022,wang2017residual,lu2007survey}, object detection \cite{QiangInstant,zhao2019object,szegedy2013deep}, and image segmentation \cite{QingAdaptive2022,pham2000survey,cheng2001color}. However, the prediction of DNN models can be easily fooled by adding some perturbation on the original image, revealing the vulnerability of the DNN models. The perturbed images are called adversarial example (AE), which can be generated by various adversarial attack algorithms \cite{goodfellow2014explaining,pgd2018}. Designing the stronger adversarial attack algorithm to evaluate the robustness of models has become an important research area. 

Recently, much effort has been devoted to improving the robustness of DNN through various defense strategies such as adversarial training, defensive distillation \cite{papernot2016distillation}, dimensionality reduction \cite{bhagoji2017dimensionality}, input transformations \cite{guo2017countering}, and activation transformations \cite{dhillon2018stochastic}. The development of the defense community makes the robustness of the DNN models difficult to be evaluated due to the weak performance of existing  adversarial attack algorithms. On the one hand, the robustness evaluation of the models using adversarial attack needs to take high calculation cost. On the other hand, the adversarial attack could not approach the lower enough bound of the robust accuracy of the defensed model.

To realize the more reliable robustness evaluation of the defensed model, some work about the auto adversarial attack has been developed. Francesco et al. \cite{ReliableFrancesco} proposed autoattack (AA) , which is the ensemble attack using four types of attack, obtaining the lower robust accuracy on multiple defense methods. Tramer et al. \cite{tramer2020adaptive} explored the influence of the loss functions on the performance of adversarial attacks, which verified that the suitable loss functions could further lower the robust accuracy of the defensed models. Mao et al. \cite{Composite2021} introduced composite adversarial attack (CAA), which utilized the NSGA-II algorithm \cite{deb2002fast} to search for near-optimal adversarial attacks considering the complexity  and robust accuracy. Tsai et al. \cite{tsai2021generalizing} extended CAA to unrestricted adversarial attack and proposed the composite adversarial training method, which can improve the robustness of models against CAA. Further, Yao \cite{Automated2021} proposed the adaptive autoattack, which searched not only the adversarial attack algorithms but also more hyperparameters such as randomization using the hyperparameters optimization search method. Liu et al. \cite{liu2022practical} proposed adpaptive adversarial attack by automatically selecting the restart direction and the attack budget on each image. In general, it can be concluded that realizing the auto adversarial attack for each defense model can obtain lower robust accuracy and provide a more reliable robustness evaluation.

However, though the above work about auto adversarial attack can obtain better attack performance than the manually-designed ones, there still exists a large space to be improved. First, the mathematical model of auto adversarial attack is not general enough. For instance, AA \cite{ReliableFrancesco}, adaptive autoattack \cite{Automated2021} and CAA \cite{Composite2021} all utilized  the ensemble of multiple adversarial attacks, which can be described as the attack sequence consisting of multiple attack cells. On the one hand, in AA and adaptive autoattack, each attack cell in the attack sequence would attack the original images that are not perturbed successfully by the previous attack cell. While in CAA, each attack cell in the attack sequence would attack the perturbed images that are not perturbed successfully by the previous attack cell.  It brings out one question: how to select the optimal connection between two adjacent attack cells in one attack sequence. On the other hand, AA did not take the time cost of robustness evaluation into consideration. Adaptive autoattack considered the time cost as a constraint in optimization, while CAA utilized the complexity of adversarial attack as an indirect metric. It is necessary to take the time cost of robustness evaluation into consideration directly. Second, the algorithms to solve the mathematical model of auto adversarial attack optimization  are not efficient enough, including obtaining better solutions and decreasing the optimization cost. For example, the NSGA-II adopted by CAA can not handle some discrete search spaces  well, leading to being easily trapped in the local optimum. In addition, the evaluation cost during the optimization would increase with the size of the dataset.

To further alleviate these problems, we propose a multi-objective memetic algorithm for auto adversarial attack optimization design. First, we propose a more general mathematical model of auto adversarial attack. The optimization objective consists of the robust accuracy of the defensed model and the time cost of evaluation by the adversarial attack. The design variable includes the attack operations, attack magnitude, iteration number, loss functions ,and restart position. Second, we develop a multi-objective memetic algorithm to solve the optimization problem, which could combine the advantages of NSGA-II and local search. In our defined auto adversarial attack optimization design problem, the proposed memetic algorithm could obtain better solutions than pure NSGA-II or local search. Third, we propose the representative data selection strategy to decrease the search cost greatly. Due to the attack transferability, we utilize the small dataset to evaluate the performance of adversarial attacks at the early stage of the optimization process, which is selected by sorting cross entropy loss values of each image.

Our contributions can be summarized as follows:

1) We propose a general mathematical model of auto adversarial attack optimization design, the search space of which includes not only the attack operation, attack loss, attack magnitude, and iteration number of each attack cell but also the connection way of different cells.

2) We propose a novel memetic algorithm to solve the auto adversarial attack optimization design problem, which combines the advantages of NSGA-II and local search, realizing the purpose of obtaining the solutions with lower robust accuracy and lower time cost efficiently.

3) We propose a representative data selection strategy to substitute the accurate evaluation in primary optimization, which selects few data from the whole dataset based on the sorting of loss values of each image, decreasing the optimization cost greatly.

The remainder of this paper is organized as follows. Section \ref{sec2} describes the preliminary knowledge about adversarial attack and defense methods. Section \ref{sec3} introduces our proposed method. Section \ref{sec4} elaborates the experimental settings and results. Section \ref{sec5} concludes this paper.

\section{Backgroud}
\label{sec2}
\begin{figure}[h]
	\centering
	\includegraphics[scale=0.4]{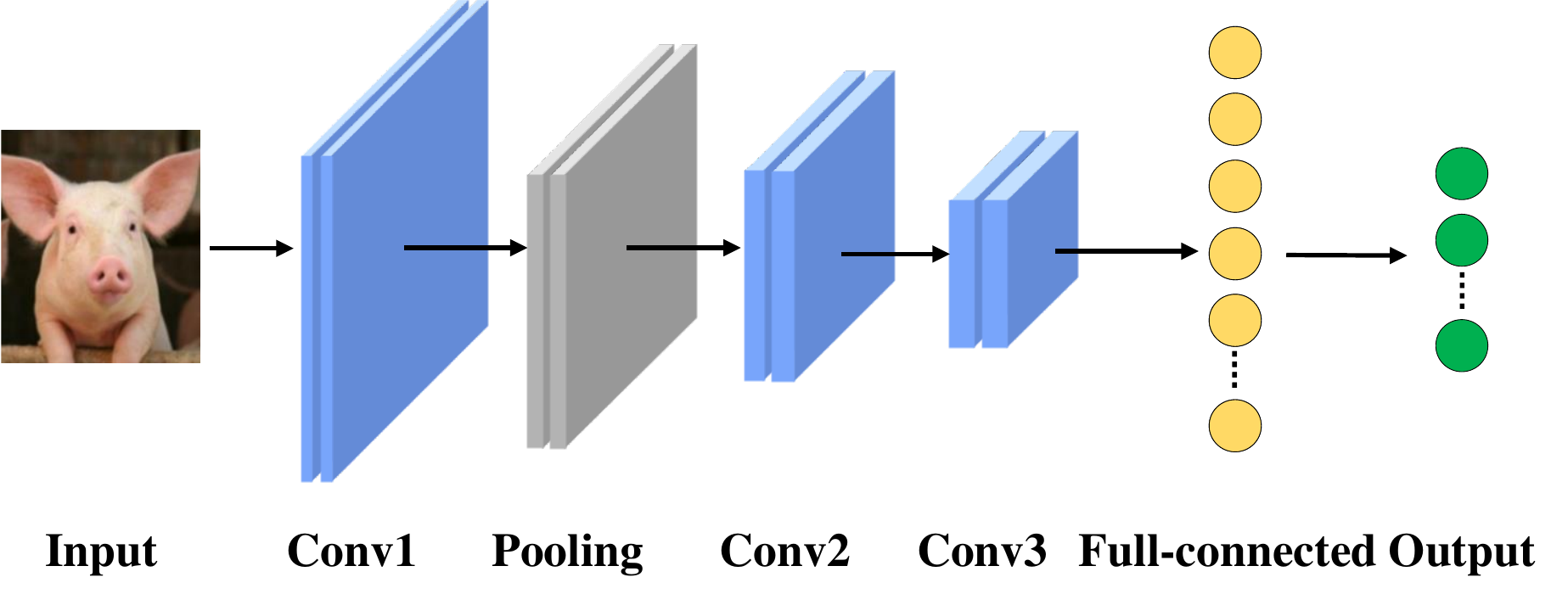}
	\caption{The process of the image classification in neural networks.}
	\label{cnn}
\end{figure}

Due to the strong ability to extract the features, deep neural networks have become the core module in numerous computer vision tasks. Taking the image classification task as an example, as presented in Fig.~\ref{cnn}, the whole neural network is composed of convolutional layers, pooling layers, and the full-connected layer. Given an input image, the output of the neural network is the prediction probability of all classes. The class with the highest probability represents the result of neural network prediction. The neural network can be trained on the dataset to obtain a satisfying enough classification accuracy.

\subsection{Adversarial attack}

\begin{figure}[h]
	\centering
	\includegraphics[scale=0.4]{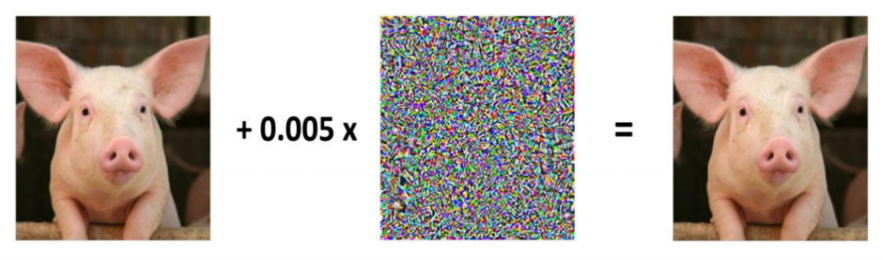}
	\caption{The illustration of adversarial attack.}
	\label{advattack}
\end{figure}

The adversarial attack could fool the trained DNN models by adding the specific perturbation on the original image. As shown in Fig.~\ref{advattack}, though the added perturbation remains imperceptible to human eyes, it can make DNN output wrong predictions. The aim of the adversarial attack is to study how to generate the perturbation to fool DNN models. Denote the original image $x$, the perturbation $\Delta x$,  the perturbation magnitude $\epsilon$, the DNN model $\mathcal{F}$, the true label $y$, the final adversarial example $x_{adv}$, the whole formula of the adversarial attack can be expressed as Eq. \ref{attack}.
\begin{equation}
\begin{array}{l}
\underset{\Delta x}{\arg \max }  \mathcal{L}\left(x_{adv}, y ; \mathcal{F}\right)\\
x_{a d v}=x+\Delta x \\
\text { s.t. }\|\Delta x\|_{p} \leq \varepsilon.
\end{array}
\label{attack}
\end{equation}
where $p$ stands for the norm type of the perturbation such as $l_{0}$, $l_{2}$ and $l_{\infty}$. $\mathcal{L}$ is the loss function.

To solve Eq. \ref{attack}, a majority of adversarial attack algorithms have been developed.

Goodfellow et al. \cite{goodfellow2014explaining} proposed fast gradient sign method (FGSM), which is a single-step updating method, as shown in Eq. \ref{fgsm}.

\begin{equation}
x_{a d v}=\operatorname{clip}_{[0,1]}\left\{x+\epsilon \cdot \operatorname{sign}\left(\nabla_{x} \mathcal{L}(x, y ; \mathcal{F})\right)\right\}
\label{fgsm}
\end{equation}

Madry et al. \cite{pgd2018} further proposed projected gradient descent (PGD), which is illustrated in Eq. \ref{pgd}.
\begin{equation}
x_{l+1}=\operatorname{project}\left\{x_{l}+\epsilon_{step} \cdot \operatorname{sign}\left(\nabla_{x} \mathcal{L}\left(x_{l}, y ; \mathcal{F}\right)\right)\right\}.
\label{pgd}
\end{equation}
where the initial image can be original image or the perturbed image by random initialization.$\epsilon_{step}$ stands for the perturbation magnitude at each iteration.

Carlini and Wagner \cite{carlini2017towards} improved the loss function of adversarial attack, which adopts the distance of logit values between the label $y$ and the second most-likely class, as presented in Eq. \ref{cw}.

\begin{equation}
\mathcal{L}_{C W}(x, y ; \mathcal{F})=\max \left(\max _{i \neq y}\left(\mathcal{F}(x)_{(i)}\right)-\mathcal{F}(x)_{(y)},-\kappa\right)
\label{cw}
\end{equation}
where $\kappa$ establishes the margin of the logit of the adversarial class being larger than the logit of runner-up class \cite{Composite2021}.

Apart from these adversarial attacks, other work such as MultiTargeted (MT) attack \cite{gowal2019alternative}, Momentum Iterative
(MI) attack \cite{dong2018boosting}, and Decoupled Direction and Norm (DDN) attack \cite{rony2019decoupling} are also further developed. To this end, it can be concluded that existing manually-designed adversarial attack methods needs to preset the magnitude of perturbation, attack iteration number, and the loss functions.

\subsection{Adversarial defense}
To improve the robustness on the adversarial examples generated by the adversarial attack, various defense methods are proposed. Adversarial training is one of the most notable defense methods, which was first introduced by Goodfellow et al. \cite{goodfellow2014explaining}. It can be modeled into a dual optimization problem consisting of the generation of adversarial examples and the trainging based on the adversarial examples. Besides, much effort has been devoted to developing other defense methods such as the defensive distillation \cite{papernot2016distillation}, dimensionality reduction \cite{bhagoji2017dimensionality}, input transformations \cite{guo2017countering} and activation transformations \cite{dhillon2018stochastic}.

With the rapid development of the defense community, it becomes more difficult to evaluate the robustness of DNN models through adversarial attacks. Given the defensed model, automatically searching the near-optimal adversarial attack that could obtain lower robust accracy with less time cost is needed to be realized.

\section{Method}
\label{sec3}
In this section, the proposed multi-objective memetic algorithm for auto adversarial attack optimization design is introduced. First, in section \ref{aaaml}, the mathematical  model of auto adversarial attack optimization design is constructed. Second, in section \ref{mma}, the multi-objective memetic algorithm for solving the optimization problem is presented in detail, which includes the search space, performance evaluation and search strategy.

\subsection{The mathematical model of auto adversarial attack}
\label{aaaml}
In this section, the multi-objective optimization mathematical model of the auto adversarial attack is constructed. Considering two objectives, including the robust accuracy and the time cost, we aim to search for the near-optimal adversarial attack sequences. An attack sequence could be determined by different attack cells, which consist of the attacker operation, attacker iteration number, attacker magnitude, attacker loss functions, and attacker restart. The mathematical model is expressed as follows: 
%\end{itemize} 

\begin{equation}
\begin{array}{ll}
\min & F(\textbf{x})=\left[f_{1}(\textbf{x}), f_{2}(\textbf{x})\right]^{T} \\
\text { s.t. } & 1 \leq \textbf{x}(5 k-4)) \leq n_{\text {attacker }} \\
& 1 \leq \textbf{x}(5 k-3)) \leq n_{\text {loss }} \\
& 1 \leq \textbf{x}(5 k-2) \leq n_{e p s} \\
& 1 \leq \textbf{x}(5 k-1) \leq n_{\text {step }} \\
& 0 \leq \textbf{x}(5 k) \leq c-1 \\
& k=1,2, \cdots, c
\end{array}
\end{equation}
where $c$ denotes the maximum number of the attack cell in one attack sequence,  $n_{\text {attacker }}$ stands for the number of the attacker operation in the search space, $n_{\text{loss}}$ represents the number of the loss functions in the search space,  $n_{\text{eps}}$ denotes the number of the attacker magnitudes in the search space, $n_{\text{step}}$ denotes the number of the attacker iteration number in the search space. $f_{1}$ is the robust accuracy of the defensed model, which is expressed as 
\begin{equation}
Robust\text{ }Accuracy\text{ =}\frac{{{n}_{adv}}}{{{n}_{total}}},
\label{eqs}
\end{equation}
where ${n}_{total}$ stands for the number of total test samples that are generated by the adversarial attack. ${n}_{adv}$ denotes the number of examples that are predicted rightly. $f_{2}$ is the time cost of the evaluation of an attack sequence on total test samples.

\begin{figure*}[h]
	\centering
	\includegraphics[scale=0.35]{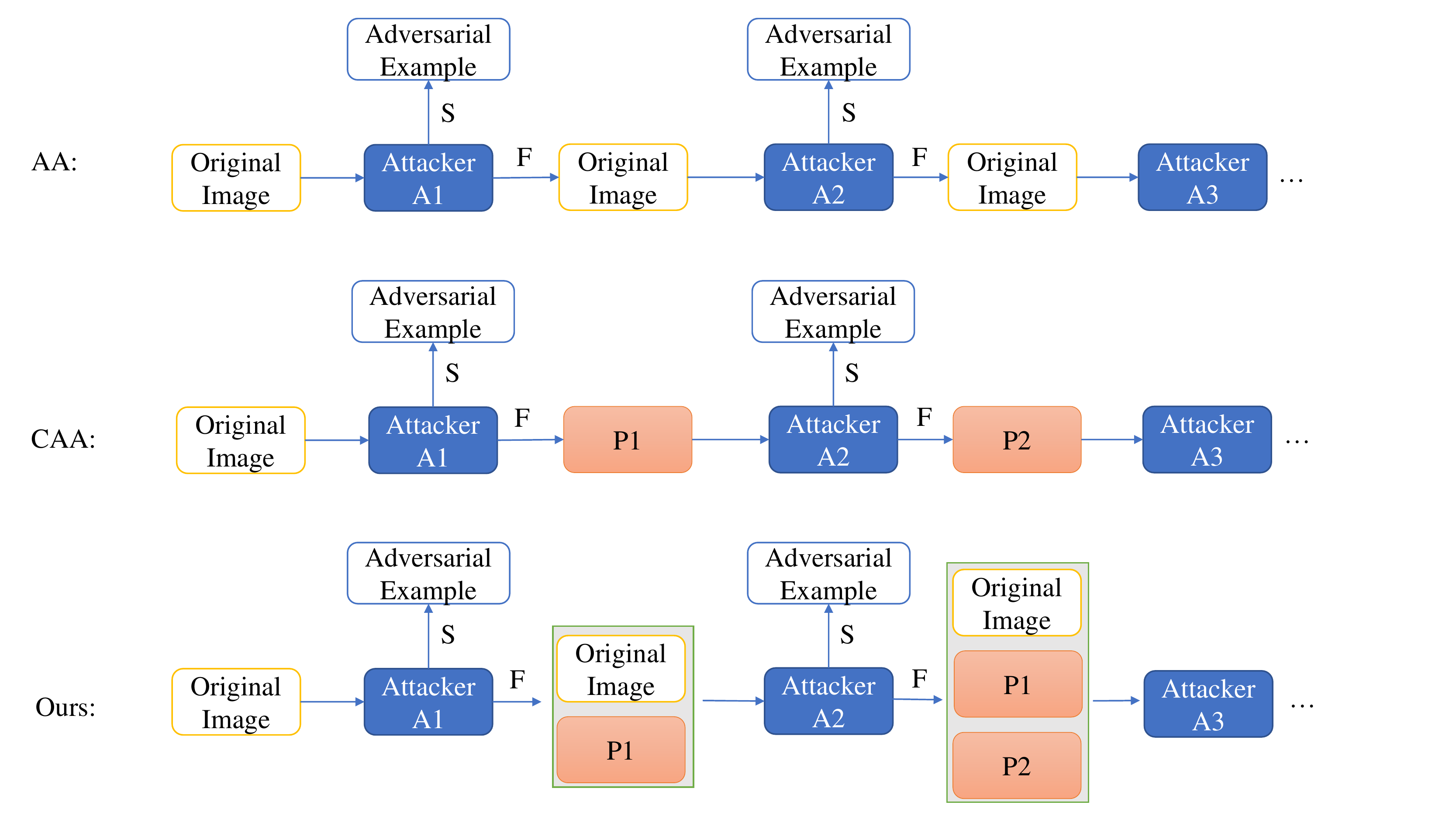}
	\caption{The illustration of the search space of auto adversarial attack.}
	\label{vcaa}
\end{figure*}

The most primary difference between the mathematical model constructed by us and the literature is illustrated in Fig.~\ref{vcaa}. In previous work, AA ensembled four different attacks in the way that the subsequent attack would perturb the original image that could not be perturbed successfully by the previous one. The original image is the input of the attack sequence. If the first adversarial attack perturbs the original image successfully, the attack process on this image terminates. If the first adversarial attack \textbf{A1} fails to perturb the original image, the original image would be input to the second attack \textbf{A2}. This process is conducted until the image is to be perturbed successfully or perturbed by all attack cells in the attack sequence. Differently, CAA searched the near-optimal attack, where the subsequent attack would continue to add the perturbation on the perturbed image by the previous attack. The output of \textbf{A1} is denoted as \textbf{P1}. The second attack \textbf{A2} would take \textbf{P1} as the input. The same operation is performed on the subsequent attacks. Both of the adversarial examples generated by AA and CAA must meet the constraint that the perturbation does not exceed the predefined maximum value. The detailed strategy to control the perturbation magnitude can be seen in \cite{Composite2021}. In our work, apart from searching the attack magnitudes, attack operations, and iteration number, we first take the connection way of different adversarial attacks into consideration. After the number of attack cells in one attack sequence is determined, all possible configurations of different connection ways are obtained, which correspond to the attacker restart.

An attack sequence example with $c$ of two is taken to illustrate the search space: {'A': PGD, 'L':CE,'M': 0.031,'I': 100, 'R': 0; 'A': FGSM,'L':CE, 'M': 0.031, 'I': 1, 'R': 0}. 'A','L','M','I','R' stand for the attack operation, attack loss function, attack magnitude, attack iteration number, and the attack restart respectively. The value of restart is selected from 0 to $c-1$, which represents the connection way between two adjacent attack. If the value of restart is 0, the original image would be as the input of the current attack. If that is 1, the perturbed image by the first attack would be as the input of the current attack. The subsequent attacks would be conducted in a similar way.

\subsection{Multi-objective memetic algorithm}
\label{mma}

\begin{figure*}[h]
	\centering
	\includegraphics[scale=0.73]{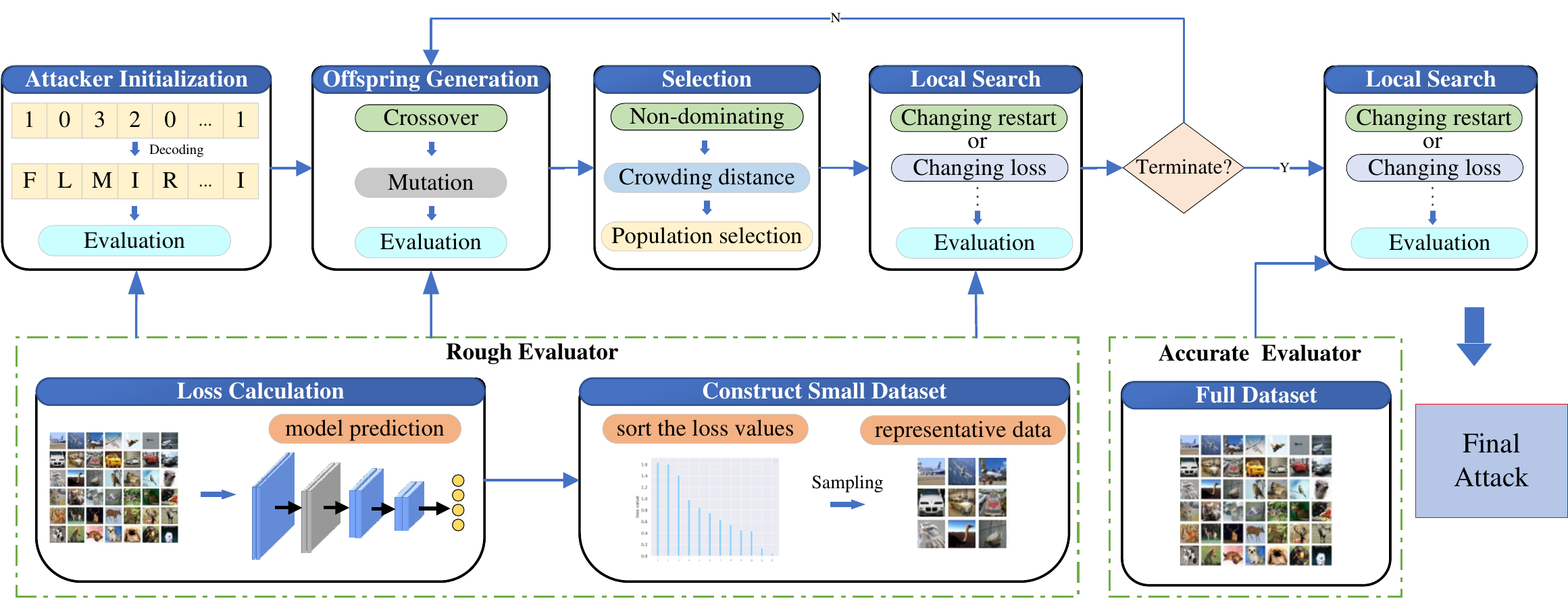}
	\caption{The illustration of the process of multi-objective memetic algorithm for auto adversarial attack.}
	\label{framework}
\end{figure*}

To solve the optimization problem introduced in section \ref{aaaml}, we propose a multi-objective memetic algorithm, which is presented in Fig.~\ref{framework}. It mainly consists of two parts. In the first part, given the defined search space, we combine the NSGA-II and local search to find a relatively optimal adversarial attack using the rough evaluator. In the second part, we further conduct local search on the selected adversarial attack on the accurate evaluator. In the following, the whole process is introduced in detail.   

\subsubsection{Search space}
Our definition of the search space includes the attacker operation, attacker loss functions, attacker magnitude, attacker iteration number, and attacker restart. The detailed search space is introduced in this section.

The search space of the attacker operations is illustrated in Table~\ref{attacker}. Two types of adversarial attacks, including the $l_{2}$ and $l_{\infty}$ norm are provided. The 
$l_{2}$ norm attacks include five attacker operations, namely FGSM-L2Attack,  PGD-L2Attack, DDNL2Attack, MT-L2Attack , and MI-L2Attack. The 
$l_{\infty}$ norm attacks include six attacker operations, namely FGSM-LinfAttack, PGD-LinfAttack, CW-LinfAttack, MT-LinfAttack , and MI-LinfAttack.

\begin{table}[h]
	\centering
	\caption{The search space of attacker operations.}
	\renewcommand{\arraystretch}{1} 
	\begin{tabular}{ccc}
		\midrule
		$l_{2}$ & \textbf{$l_{\infty}$}  \\
		\midrule
		DDNL2Attack &  FGSM-LinfAttack\\
		PGD-L2Attack  & PGD-LinfAttack\\
		CW-L2Attack & CW-LinfAttack \\
		MT-L2Attack    & MT-LinfAttack   \\
		MI-L2Attack      & MI-LinfAttack   \\
%		& Momentum-LinfAttack\\
		\midrule
	\end{tabular}
	\label{attacker}
\end{table}
Besides, we provide different attacker loss functions, which are illustrated in Fig.~\ref{caa}. It has been proved that the loss functions have a great influence on the performance of adversarial attacks. We include four type of loss functions: $l_{\text{CE}}$, $l_{\text{Hinge}}$, $l_{\text{L1}}$ and $l_{\text{DLR}}$. Further, calculating the output of logit or probability is also one key factor for the performance of adversarial attacks. Logit stands for the output of the final full-connected layer of DNNs, while probability means the output handled by the softmax operation. Due to the calculation of cross entropy needs to ensure the positive input, the logit output is not considered. Thus, the search space of provided loss functions can form into seven loss functions totally. In practical, due to the specific design of the loss function of CW-Attack itself, we do not take the loss functions of that into consideration. 
\begin{table}[h]
	\centering
	\caption{The search space of attacker loss functions.}
	\renewcommand{\arraystretch}{1} 
	\begin{tabular}{cccc}
		\midrule
		\multicolumn{1}{c}{Name} &Logit/Prob & Loss  \\
		\midrule
		\multicolumn{1}{c}{$l_{\text{CE}}$}    & P&  $
		-\sum_{i=1}^{K} y_{i} \log \left(Z(x)_{i}\right)
		$\\
		\multicolumn{1}{c}{$l_{\text{Hinge}}$}     &L/P & $
		\max \left(-Z(x)_{y}+\max _{i \neq y} Z(x)_{i},-\kappa\right)
		$ \\
		\multicolumn{1}{c}{$l_{\text{L1}}$}     &L/P & $
		-Z(x)_{y}
		$ \\
		\multicolumn{1}{c}{$l_{\text{DLR}}$}     &L/P & $
		-\frac{Z(x)_{y}-\max _{i \neq y} Z(x)_{i}}{Z(x)_{\pi_{1}}-Z(x)_{\pi_{3}}}
		$   \\
		\midrule
	\end{tabular}
	\label{caa}
\end{table}

Further, in our defined search space, the iteration number and magnitude are divided unfiormly into 8 parts. Thus, each attacker sequence consists of multiple attack cells. Given the predefined maximum magnitude, the magnitude of the generated adversarial example by the attacker sequence is required to not exceeed it. 

\subsubsection{Performance evaluation}
The evaluation of the performance of adversarial attack is time-consuming. If each adversarial attack is directly evaluated on the whole dataset, it would bring huge computational cost during optimization. 

To accelerate the optimization, we propose the way of combining the rough evaluator and accurate evaluator together. First, in early optimization, the rough evaluator with the small dataset is utilized to fastly find a relatively optimal solution. Second, in later optimization, the accurate evaluator with the whole dataset is adopted to improve the performance of the found adversarial attack. In detail, given the whole dataset, the representative data is needed to be selected to form the small dataset. We first calculate the loss values of each image and sort them in order. Then the small dataset is obtained by uniformly sampling in the sorted order. The reason for selecting the loss values as the metric to sample out a small dataset is that the loss values could reflect the degree of being attacked successfully for each image. We hope that we can select more diverse images other than all the easily-attacked or hardly-attacked images. Compared with the random selection strategy, our proposed representative data selection strategy could distinguish the performance of different adversarial attacks better.

\subsubsection{Search strategy}
To solve the above multi-objective optimization problem, the memetic algorithm that combines NSGA-II and local search is introduced.

In the population initialization, we generate the initial population $\bm{P}$  with $n$ candidate adversarial attack sequences randomly. The length of each adversarial attack does not exceed $c$. Compared with the random generation of the magnitude in each attacker cell, we set all magnitudes of the attack cells to the predefined maximum magnitude. It is because if none of the magnitude of the attacker cell in the sequence reaches the predefined maximum magnitude, the performance of the attacker sequence would be farther worse than the expectation. In addition, it would make the optimization algorithm difficult to obtain the near-optimal adversarial attack.

\begin{algorithm}[h]
	\caption{Multi-objective Memetic Algorithm based Auto Adversarial Attack}
	\label{alg3}
	\LinesNumbered
	\KwIn{\\ Input: The maximum number of iteration ${T}$ in NSGA-II, the maximum number of iteration ${T_{l}}$ in local search, population size $n$, training set $D_{train}$, validation set $D_{test}$, the crossover rate $P_{c}$, the mutation rate $P_{m}$, the maximum length of the searched adversarial attack $c$;}
	\KwOut{\\ The near optimal adversarial attack.}
	Generate the initial population $\bm{P}$  with $n$ candidate adversairial attack randomly. The length of each adversarial attack does not exceed $c$;
	\\
	Calculate the loss values of each image in the whole dataset $D_{train}$ using the model and sort them in descend order;\\
	Obtain the \textbf{rough evaluator} by uniform sampling technique;\\ 
	Evaluate the robustness accuracy and time cost of each individual in  $\bm{P}$;\\
	
	\While{$t \leftarrow 1:T$}{
		\textbf{// Evolution by discrete NSGA-II using rough evaluator}\\
		Generate $n$ offspring individuals using crossover and mutation operation;\\
		Merge the offspring individuals with $\bm{P}$ and conduct the evaluation;\\
		Select $n$ individuals to form the new $\bm{P}$ using non-dominated-sorting and crowding distance;\\ 
		\textbf{// Local search using rough evaluator}
		Obtain the individual in the first Pareto  $\bm{P}_{best}$;\\
		Randomly select the local search operation, including changing the restart, loss function, magnitude, iteration number and attacker operation based on $\bm{P}_{best}$;\\
		Evaluate the individuals generated by local search and select the best one to substitute $\bm{P}_{best}$;\\	
	}
	\textbf{// Local search using accurate evaluator}\\
	Select $l$ individuals with different lengths;\\
	Evaluate these individuals using \textbf{accurate evaluator} on the whole dataset and select the best one;\\
	Conduct local search based on the best adversarial attack using Algorithm \ref{alg4};\\
	
	Return The near optimal adversarial attack
\end{algorithm}

Evolutionary algorithms need to take a large number of evaluations, which would bring out the huge calculation cost if each adversarial attack is evaluated in the whole dataset. Thus, we adopt the two-stage strategy. In the first stage, a rough evaluator obtained by sampling a small dataset from the whole one is utilized. In the second stage, an accurate evaluator of the whole dataset would be utilized to evaluate the performance of adversarial attacks.

	In detail, our proposed search strategy is as follows. In the first stage, we combine the NSGA-II and local search together. The initial population is generated by randomly sampling different adversarial attack sequences. Among that, to accelerate the process of optimization, the attack magnitudes of attack cells in all attack sequences are set to the pre-defined maximum value. The initial population can be regarded as the parent population. Following that, we conduct the classical crossover, mutation, and selection operations to generate the children population. Each individual in the children population is generated by the crossover and mutation operation on two individuals selected by tournament selection strategy. Further, the individual of the first pareto is randomly selected to conduct the local search, improving the optimization performance, where the neighbor space is constructed by randomly changing the attacker magnitude, attacker restart, and attacker iteration number at each time. At each time, we just change one element in one attack cell. For instance, as for one attack sequence  with length of three, if the attacker restart is changed, the neighbor space has five individuals totally. If the attacker loss function is changed, the neighbor space has twenty individuals totally. These individuals generated by local search would be evaluated and compared with the selected one. If the local search finds a better individual, the found one would substitute the original one, regarded as one of the parent population in the next iteration. In the second stage, the detailed procedure of local search on the accurate evaluator is shown in Algorithm \ref{alg4}. We select $l$ individuals with different lengths and select the best one on the accurate evaluator, which can prevent being trapped in the local optimum. The local search is performed on the best individual using the accurate evaluator. During the local search, the solution with the lowest robust accuracy is regarded as the best individual. If there exist multiple solutions with the same robust accuracy, the solution with the least time cost is regarded as the best individual.

\begin{algorithm}[h]
	\caption{Local search based on one adversarial attack}
	\label{alg4}
	\LinesNumbered
	\KwIn{\\ Input: the maximum number of iteration ${T_{l}}$ in local search, initial attack $\textbf{x}$, training set $D_{train}$;}
	\KwOut{\\ The near optimal adversarial attack $\textbf{x}^{*}$.}
	The best attack $\textbf{x}^{*}$ $\leftarrow$ initial attack $\textbf{x}$\\
	\While{$t \leftarrow 1:T_{l}$}{
		Generate one inter sequence from [1,4] randomly;\\
		\For{$k \leftarrow 1:4$}{
			\If{$k==1$}{
				Generate the nerighbor space by changing the restart of $\textbf{x}^{*}$;\\
			}
			\If{$k==2$}{
				Generate the nerighbor space by changing the loss functions of $\textbf{x}^{*}$;
			}
			\If{$k==3$}{
				Generate the nerighbor space by changing the length of $\textbf{x}^{*}$;\\
			}
			\If{$k==4$}{
				Generate the nerighbor space by changing the attacker operation of $\textbf{x}^{*}$;\\
			}
			Evaluate all the individuals in the neighbor space;\\ 
			The best attack $\textbf{x}^{*}$$\leftarrow$ The best individual in the neighbor space; \\
		}
	}	
	Return The near optimal adversarial attack $\textbf{x}^{*}$
\end{algorithm}

\section{Experiments}
\label{sec4}
\subsection{Experiment Protocol}
To verify the necessities and effectiveness of our proposed AAA, we conduct experiments on CIFAR10, CIFAR100 and ImageNet dataset for image classification task.  

\textbf{Searching for adversarial attack}: In the process of searching the efficient adversarial attack, we limit the maximum length of the attacker sequence to four. The maximum iteration number of each attack operation is set to 200. To accerlerate the attack evaluation, the number of training samples in the rough evaluator is set to 36. The maximum magnitudes in $l_{\infty}$ and $l_{2}$ attack for CIFAR10 classification task are set to 0.031 and 0.5, respectively, and the maximum magnitudes in $l_{\infty}$ attack for ImageNet classification task is set to 4/255. In the NSGA-II, we set the number of the generation to 20. In each generation, the population size is kept to 40. The crossover rate and mutation rate are set to 0.8 and 0.6 respectively. In the local search, $l$ and $T_{l}$ are set to three and one, respectively. Various defensed models are selected from \cite{liu2022practical}.

\textbf{Evaluation}: In the process of evaluating the performance of different adversarial attack, we select 500, 500, 400 test images from CIFAR10, CIFAR100 and ImageNet dataset respectively. The comparison methods include APGD \cite{ReliableFrancesco}, AA \cite{ReliableFrancesco} and CAA \cite{Composite2021}. APGD and AA are the manually-designed adversarial attack with superior performance, while CAA is the recent search based adversarial attack.

The above experiments are conducted under the same environment using a single NVIDIA GTX 3090Ti GPU.

\setlength{\tabcolsep}{1.4pt}

\begin{table*}[t]
	\renewcommand\arraystretch{1.5}
	\scriptsize
	\centering
	\caption{Comparison of the robust accuracy ($\%$) and time cost (min) under the $l_{\infty}$  attack of APGD, AA and AAA across various defense strategies with the magnitude set as 0.031 using 500 images on CIFAR10 dataset.}
	
	\label{cifar10linf}
	\setlength{\tabcolsep}{4mm}{
		\begin{tabular}{c|c|cccccccccc}
			
			\toprule
			
			\multirow{2}{*}{Defense Method}&	\multirow{2}{*}{Model} &\multicolumn{1}{c}{Clean}& \multicolumn{1}{c}{APGD}& \multicolumn{2}{c}{AA} &  \multicolumn{2}{c}{AAA$_{sub}$(Ours)} &\multicolumn{2}{c}{AAA(Ours)} \\
			
			\cmidrule(r){3-3}  \cmidrule(r){4-4}  \cmidrule(r){5-6}  \cmidrule(r){7-8}
			\cmidrule(r){9-10}
			&	& acc & acc	 &   acc		
			&  time cost     &  acc   &   time cost &  acc   &   time cost \\		
			\midrule		
			{PGDAT \cite{pgd2018}}&	{ResNet-50}   &   89.6  &    52.80   &50.8 &90.98& \textbf{50.6} & \textbf{10.13} &   \textbf{50.6} & \textbf{10.13}  \\	
			%			{ULAT \cite{gowal2020uncovering}}&	{WRN-70-16}   &                     \\	
			%						{Fixing Data \cite{rebuffi2021fixing}}	&	{WRN-70-16}            \\	
			%						{ULAT \cite{gowal2020uncovering}}	&	{WRN-28-10}                   \\	
			%					 	{Fixing Data \cite{rebuffi2021fixing}}	&	{WRN-28-10}                     \\
			%						{RLPE \cite{sridhar2021robust}}&	\text { WRN-34-15 }   \\
			%		{Geomoetry}	&	\text {WRN-28-10 }        \\
			{AWP \cite{wu2020adversarial}}	&	\text {WRN-28-10 }     &89.8 &61.8 &\textbf{59.0} &227.11 &59.2 & 19.33  &\textbf{59.0} & \textbf{19.13} \\
			{AWP \cite{wu2020adversarial}}	&	\text {WRN-34-10 }     &85.8 &57.2 &\textbf{54.8} &130.25 & \textbf{54.8} &   22.88 & \textbf{54.8} &   \textbf{15.88} \\
			%			{RST}	&	\text { WRN-28-10 }                       \\
			
			%			{Proxy}&	{WRN-34-10}   &         \\	
			%				{OAAT} &    WRN-34-10  &            \\	
			%			HYDRA &	{WRN-28-10}                                                        \\	
			%						ULAT&	{WRN-70-16}                  \\
			
			%						ULAT&	\text { WRN-34-20} \\
			MART \cite{wang2019improving} & WRN-28-10 &88.4 & 62.0 &56.6 &123.96 & 56.6&  19.10&\textbf{56.4}&\textbf{17.12} \\
			%			Pre-training & WRN-34-10\\
			%			{Proxy}& ResNet-18\\
			%			AT$\_$HE &WRN-34-20\\
			%			LBGAT &WRN-34-20\\
			%			FAT &WRN-34-10\\
			%			Overfitting &WRN-34-20\\
			%			Self-adaptive &WRN-34-10\\
			TRADES \cite{zhang2019theoretically} & WRN-34-10 & 86.8 &  53.80 &\textbf{51.6} &122.61 &  \textbf{51.6} & 22&\textbf{51.6} & \textbf{18.07}\\
			%			LBGAT & WRN-34-10\\
			%			OAAT & ResNet-18\\
			%			SAT & WRN-34-10\\
			%			Robustness &ResNet-50\\
			%			YOPO & WRN-34-10\\
			%			MMA & WRN-28-4\\
			%			DNR & ResNet-18\\
			%			CNL & ResNet-18\\
			Feature Scatter \cite{zhang2019defense} & WRN-28-10 & 92.0  &62.8 & 37.2 &166.07 & 36.8 & \textbf{15.99}& \textbf{36.6} &16.58\\
			Interpolation \cite{zhang2019adversarial} & WRN-28-10 &90.6 &70.6 &\textbf{35.4} &81.63 &35.6 &  16.05 &\textbf{35.4} &\textbf{15.85} \\
			%			Sensible &WRN-34-10\\
			Regularization \cite{jin2020manifold} & ResNet-18 & 92.2 & 54.8& 6 &8.53 & 3.6 &  \textbf{4.99} &\textbf{1.8} & 11.48\\

			\bottomrule		
	\end{tabular}}
\end{table*}

\begin{table*}[t]
	\renewcommand\arraystretch{1.5}
	\scriptsize
	\centering
	\caption{Comparison of the robust accuracy ($\%$) and time cost (min) under the $l_{2}$  attack of APGD, AA and AAA across various defense strategies with the magnitude set as 0.5 using 500 images on CIFAR10 dataset.}
	
	\label{cifar10l2}
	\setlength{\tabcolsep}{4mm}{
		\begin{tabular}{c|c|cccccccccc}
			
			\toprule
			
			\multirow{2}{*}{Defense Method}&	\multirow{2}{*}{Model} &\multicolumn{1}{c}{Clean}& \multicolumn{1}{c}{APGD}& \multicolumn{2}{c}{AA} &  \multicolumn{2}{c}{AAA$_{sub}$(Ours)} &\multicolumn{2}{c}{AAA(Ours)} \\
			
			\cmidrule(r){3-3}  \cmidrule(r){4-4}  \cmidrule(r){5-6}  \cmidrule(r){7-8}
			\cmidrule(r){9-10}
			&	& acc & acc	 &   acc		
			&  time cost     &  acc   &   time cost &  acc   &   time cost \\		
			\midrule		
			{PGDAT \cite{pgd2018}}&	{ResNet-50}   &   89.6  &   61.8  &57.6 & 175.65&\textbf{57.2}&\textbf{16.08}&\textbf{57.2} &\textbf{16.08}\\	
			%			{ULAT \cite{gowal2020uncovering}}&	{WRN-70-16}   &                     \\	
			%			{Fixing Data \cite{rebuffi2021fixing}}	&	{WRN-70-16}            \\	
			%			{ULAT \cite{gowal2020uncovering}}	&	{WRN-28-10}                   \\	
			%		 	{Fixing Data \cite{rebuffi2021fixing}}	&	{WRN-28-10}                     \\
			%			{RLPE \cite{sridhar2021robust}}&	\text { WRN-34-15 }   \\
			%		{Geomoetry}	&	\text {WRN-28-10 }        \\
			{AWP \cite{wu2020adversarial}}	&	\text {WRN-28-10 }     &89.8 &66.4&65.0&260.77&64.8&38.13&\textbf{64.6} &\textbf{26.2}\\
			{AWP \cite{wu2020adversarial}}	&	\text {WRN-34-10 }     &85.8 &62.0&59.8&314.12&60.0&\textbf{63.18}& \textbf{59.6} &69.23\\
			%			{RST}	&	\text { WRN-28-10 }                       \\
			
			%			{Proxy}&	{WRN-34-10}   &         \\	
			%				{OAAT} &    WRN-34-10  &            \\	
			%			HYDRA &	{WRN-28-10}                                                        \\	
			%			ULAT&	{WRN-70-16}                  \\
			
			%			ULAT&	\text { WRN-34-20} \\
			MART \cite{wang2019improving} & WRN-28-10 &88.4 &67.8&62.6&259.15&62.6&29.17 &\textbf{62.4}&\textbf{28.5}\\
			%			Pre-training & WRN-34-10\\
			%			{Proxy}& ResNet-18\\
			%			AT$\_$HE &WRN-34-20\\
			%			LBGAT &WRN-34-20\\
			%			FAT &WRN-34-10\\
			%			Overfitting &WRN-34-20\\
			%			Self-adaptive &WRN-34-10\\
			TRADES \cite{zhang2019theoretically} & WRN-34-10 & 86.8 & 59.8 & 58.4 &315.77&58.8 &36.95 &\textbf{58.2} &\textbf{34.8}\\
			%			LBGAT & WRN-34-10\\
			%			OAAT & ResNet-18\\
			%			SAT & WRN-34-10\\
			%			Robustness &ResNet-50\\
			%			YOPO & WRN-34-10\\
			%			MMA & WRN-28-4\\
			%			DNR & ResNet-18\\
			%			CNL & ResNet-18\\
			Feature Scatter \cite{zhang2019defense} & WRN-28-10 & 92.0  &71.8&52.8&264.42&53.0&40.17 &\textbf{52.0}&\textbf{39.1}\\
			Interpolation \cite{zhang2019adversarial} & WRN-28-10 &90.6&73.8&50.4&263.37 & 45.2 &\textbf{41}&\textbf{43.4} &54.28\\
			%			Sensible &WRN-34-10\\
			Regularization \cite{jin2020manifold} & ResNet-18 & 92.2& 79.2 & \textbf{20.6}&63.83&27.4 &55.18 &23.4 &\textbf{23.97}\\
			\bottomrule		
	\end{tabular}}
\end{table*}

\begin{table*}
	\renewcommand\arraystretch{1.5}
	\scriptsize
	\centering
	\caption{Comparison of the robust accuracy ($\%$) and time cost (min) under the $l_{\infty}$ attack of APGD, AA and AAA across various defense strategies with the magnitude set as 0.031 using 500 images on CIFAR100 dataset.}
	
	\label{cifar100linf}
	\setlength{\tabcolsep}{4mm}{
		\begin{tabular}{c|c|cccccccccc}
			
			\toprule
			
			\multirow{2}{*}{Defense Method}&	\multirow{2}{*}{Model} &\multicolumn{1}{c}{Clean}& \multicolumn{1}{c}{APGD}& \multicolumn{2}{c}{AA} &  \multicolumn{2}{c}{AAA$_{sub}$(Ours)} &\multicolumn{2}{c}{AAA(Ours)} \\
			
			\cmidrule(r){3-3}  \cmidrule(r){4-4}  \cmidrule(r){5-6}  \cmidrule(r){7-8}
			\cmidrule(r){9-10}
			&	& acc & acc	 &   acc		
			&  time cost     &  acc   &   time cost &  acc   &   time cost \\		
			\midrule		
			%			{ULAT \cite{gowal2020uncovering}}&	{WRN-70-16}   &                     \\	
			{Fixing Data \cite{rebuffi2021fixing}}	&	{WRN-28-10}    &59.2 & 35.2&\textbf{31.4}   &126.22    &31.6 &13.72&\textbf{31.4}&\textbf{12.03}\\	
			{OAAT \cite{addepalli2021towards}}	&	{ResNet-18}       & 60.80&32.8& \textbf{26.8}&20.47 &27.0 &9.58 &\textbf{26.8}    &  \textbf{6.2}    \\	
			{OAAT \cite{addepalli2021towards}}	&	{WRN-34}       & 64.20&37.0& 33.0&128.44 &33.0&27.47 & \textbf{32.8}  &\textbf{21.3}\\
			{LBGAT \cite{cui2021learnable}}	&	WRN-34-10    &59.40&34.80&\textbf{30.0}&118.12&30.4 &14.98&\textbf{30.0}&\textbf{3.55} \\
			{AWP \cite{wu2020adversarial}}	&	WRN-34-10   &59.20  &33.0& \textbf{30.00}&116.70&\textbf{30} &24.92 &\textbf{30.0}& \textbf{10.32} \\
			{IAR \cite{bernhard2020luring}}	&	WRN-34-10   &61.60  &27.00& \textbf{24.8}&99.006&\textbf{24.8}&23.92&\textbf{24.8} &\textbf{11.87}\\
			{Overfit \cite{rice2020overfitting}}	&	\text {PAResNet-18 }  &53.20  &22.00& \textbf{20.6}&15.81&\textbf{20.6}&8.7&\textbf{20.6} &\textbf{2.52}\\

			\bottomrule		
	\end{tabular}}
\end{table*}

\begin{table*}
	\renewcommand\arraystretch{1.5}
	\scriptsize
	\centering
	\caption{Comparison of the robust accuracy ($\%$) and time cost (min) under the $l_{\infty}$ attack of APGD, AA and AAA across various defense strategies with the magnitude set as 4/255 using 400 images on ImageNet dataset.}
	
	\label{ImageNetlinf}
	\setlength{\tabcolsep}{4mm}{
		\begin{tabular}{c|c|cccccccccc}
			
			\toprule
			
			\multirow{2}{*}{Defense Method}&	\multirow{2}{*}{Model} &\multicolumn{1}{c}{Clean}& \multicolumn{1}{c}{APGD}& \multicolumn{2}{c}{AA} &  \multicolumn{2}{c}{AAA$_{sub}$(Ours)} &\multicolumn{2}{c}{AAA(Ours)} \\
			
			\cmidrule(r){3-3}  \cmidrule(r){4-4}  \cmidrule(r){5-6}  \cmidrule(r){7-8}
			\cmidrule(r){9-10}
			&	& acc & acc	 &   acc		
			&  time cost     &  acc   &   time cost &  acc   &   time cost \\		
			\midrule		
			%			{ULAT \cite{gowal2020uncovering}}&	{WRN-70-16}   &                     \\	
			{FastAT \cite{wong2020fast}}	&	{WRN-28-10}    &46.5 & 23.0&\textbf{21.0}   &105.88  &21.5 & \textbf{22.77} &21.25 &42.27\\	
			{Salman \cite{salman2020adversarially}}	&	{ResNet-18}       & 47.5&24.25& \textbf{22.25}&  39.77&\textbf{22.25}     &8.82     &\textbf{22.25}   &\textbf{6.23}   \\	
			{Salman \cite{salman2020adversarially}}	&	{ResNet-50}       & 57.25&32.75& 31.25&151.0 &31.25 & 28.0  &\textbf{30.88}& \textbf{14.5}\\

			\bottomrule		
	\end{tabular}}
\end{table*}

\begin{table*}
	\renewcommand\arraystretch{1.5}
	\scriptsize
	\centering
	\caption{The visualization of the searched $l_{\infty}$ attack by AAA across various defense strategies on CIFAR10 dataset. 'A' stands for the attacker operation,'L'is the loss function, 'M' denotes the attack magnitude, 'I' represents the iteration number, 'R' is the restart point.}
	
	\label{linf}
	\setlength{\tabcolsep}{4mm}{
		\begin{tabular}{ccc}		
			\toprule
			{Defense Method}&	{Model} &{The searched attack}\\
			\midrule		
			{PGDAT \cite{pgd2018}}&	{ResNet-50}   &\makecell[c]{'A': MT-LinfAttack, 'L':CE, 'M': 0.031,'I': 100, 'R': 0; 'A': FGSM-LinfAttack,'L':CE, \\'M': 0.031, 'I': 1, 'R': 0}  \\	
			\midrule
			%			{ULAT \cite{gowal2020uncovering}}&	{WRN-70-16}   &                     \\	
			%						{Fixing Data \cite{rebuffi2021fixing}}	&	{WRN-70-16}            \\	
			%						{ULAT \cite{gowal2020uncovering}}	&	{WRN-28-10}                   \\	
			%					 	{Fixing Data \cite{rebuffi2021fixing}}	&	{WRN-28-10}                     \\
			%						{RLPE \cite{sridhar2021robust}}&	\text { WRN-34-15 }   \\
			%		{Geomoetry}	&	\text {WRN-28-10 }        \\
			{AWP \cite{wu2020adversarial}}	&	\text {WRN-28-10 } & \makecell[c]{'A': MT-LinfAttack, 'L':CE, 'M': 0.031,'I': 50, 'R': 0; 'A': MT-LinfAttack,'L':DLR$\_$P, 'M': 0.031, \\'I': 25, 'R': 1; 'A': CW-LinfAttack, 'M': 0.031,'I': 125, 'R': 2}  \\
			\midrule	     
			{AWP \cite{wu2020adversarial}}	&	\text {WRN-34-10 }   & \makecell[c]{'A': CW-LinfAttack, 'M': 0.031,'I': 12, 'R': 1; 'A': FGSM-LinfAttack,\\'L':DLR$\_$P, 'M': 0.031, 'I': 1, 'R': 0; 'A': CW-LinfAttack, 'M': 0.031,'I': 12, 'R': 1}     \\
			\midrule
			%			{RST}	&	\text { WRN-28-10 }                       \\
			
			%			{Proxy}&	{WRN-34-10}   &         \\	
			%				{OAAT} &    WRN-34-10  &            \\	
			%			HYDRA &	{WRN-28-10}                                                        \\	
			%						ULAT&	{WRN-70-16}                  \\
			
			%						ULAT&	\text { WRN-34-20} \\
			MART \cite{zhang2019theoretically} & WRN-28-10  & \makecell[c]{'A': MT-LinfAttack,'L':CE, 'M': 0.031,'I': 50, 'R': 0; 'A': MT-LinfAttack,\\'L':CE, 'M': 0.031, 'I': 25, 'R': 1}     \\
			\midrule
			%			Pre-training & WRN-34-10\\
			%			{Proxy}& ResNet-18\\
			%			AT$\_$HE &WRN-34-20\\
			%			LBGAT &WRN-34-20\\
			%			FAT &WRN-34-10\\
			%			Overfitting &WRN-34-20\\
			%			Self-adaptive &WRN-34-10\\
			TRADES \cite{zhang2019theoretically} & WRN-34-10 &  \makecell[c]{'A': MT-LinfAttack, 'L':CE,'M': 0.031,'I': 50, 'R': 0; 'A': FGSM-LinfAttack,\\'L':CE, 'M': 0.031, 'I': 1, 'R': 1; 'A': CW-LinfAttack, 'M': 0.031,'I': 125, 'R': 2}     \\
			\midrule
			%			LBGAT & WRN-34-10\\
			%			OAAT & ResNet-18\\
			%			SAT & WRN-34-10\\
			%			Robustness &ResNet-50\\
			%			YOPO & WRN-34-10\\
			%			MMA & WRN-28-4\\
			%			DNR & ResNet-18\\
			%			CNL & ResNet-18\\
			Feature Scatter \cite{zhang2019defense} & WRN-28-10  & \makecell[c]{'A': CW-LinfAttack, 'M': 0.031,'I': 100, 'R': 0; 'A': FGSM-LinfAttack,\\'L':CE, 'M': 0.031, 'I': 25, 'R': 1; 'A': MT-LinfAttack, 'L':CE, 'M': 0.031,'I': 125, 'R': 0}     \\
			\midrule
			Interpolation \cite{zhang2019adversarial} & WRN-28-10  & \makecell[c]{'A': 'MT-LinfAttack', 'L': 'CE', 'M': 0.031, 'I': 50.0, 'R': 0, 'A': 'MT-LinfAttack', 'L': 'CE',\\ 'M': 0.031, 'I': 25.0, 'R': 0, 'A': 'CW-LinfAttack', 'L': 'DLR$\_$P', 'M': 0.027, 'I': 125.0, 'R': 0}     \\
			\midrule
			%			Sensible &WRN-34-10\\
			Regularization \cite{jin2020manifold} & ResNet-18  & \makecell[c]{'A': MT-LinfAttack,'L': CE, 'M': 0.031,'I': 50, 'R': 0; 'A': MT-LinfAttack,\\'L':DLR, 'M': 0.027, 'I': 25, 'R': 0; 'A': MT-LinfAttack, 'L':DLR,'M': 0.031,'I': 125, \\'R': 2;'A': MI-LinfAttack, 'L':DLR$\_$P,'M': 0.031,'I': 200, 'R': 0}     \\

			\bottomrule		
	\end{tabular}}
\end{table*}

\begin{table*}
	\renewcommand\arraystretch{1.5}
	\scriptsize
	\centering
	\caption{The visualization of the searched $l_{2}$ attack by AAA across various defense strategies on CIFAR10 dataset.}
	
	\label{l2}
	\setlength{\tabcolsep}{4mm}{
		\begin{tabular}{ccc}		
			\toprule
			{Defense Method}&	{Model} &{The searched attack}\\
			\midrule		
			{PGDAT \cite{pgd2018}}&	{ResNet-50}   &\makecell[c]{'A': MT-L2Attack, 'L':CE, 'M': 0.375,'I': 100, 'R': 0; 'A': DDNL2Attack,'L':CE, \\'M': 0.5, 'I': 25, 'R': 1; 'A': DDNL2Attack, 'L':DLR$\_$P,'M': 0.5,'I': 125, 'R': 0}  \\	
			\midrule
			%			{ULAT \cite{gowal2020uncovering}}&	{WRN-70-16}   &                     \\	
			%						{Fixing Data \cite{rebuffi2021fixing}}	&	{WRN-70-16}            \\	
			%						{ULAT \cite{gowal2020uncovering}}	&	{WRN-28-10}                   \\	
			%					 	{Fixing Data \cite{rebuffi2021fixing}}	&	{WRN-28-10}                     \\
			%						{RLPE \cite{sridhar2021robust}}&	\text { WRN-34-15 }   \\
			%		{Geomoetry}	&	\text {WRN-28-10 }        \\
			{AWP \cite{wu2020adversarial}}	&	\text {WRN-28-10 } & \makecell[c]{'A': MT-L2Attack, 'L':CE, 'M': 0.5,'I': 100, 'R': 0; \\'A': PGD-L2Attack,'L':CE, 'M': 0.5, 'I': 25, 'R': 1}  \\
			\midrule	     
			{AWP \cite{wu2020adversarial}}	&	\text {WRN-34-10 }   & \makecell[c]{'A': MI-L2Attack,'L':CE, 'M': 0.5,'I': 100, 'R': 0; 'A': PGD-L2Attack,\\'L':CE, 'M': 0.375, 'I': 25, 'R': 1; 'A': DDNL2Attack, 'L':DLR$\_$P,'M': 0.5,'I': 125, 'R': 2;\\'A': MT-L2Attack, 'L':DLR,'M': 0.5,'I': 200, 'R': 0}     \\
			\midrule
			%			{RST}	&	\text { WRN-28-10 }                       \\
			
			%			{Proxy}&	{WRN-34-10}   &         \\	
			%				{OAAT} &    WRN-34-10  &            \\	
			%			HYDRA &	{WRN-28-10}                                                        \\	
			%						ULAT&	{WRN-70-16}                  \\
			
			%						ULAT&	\text { WRN-34-20} \\
			MART \cite{zhang2019theoretically} & WRN-28-10  & \makecell[c]{'A': CW-LinfAttack, 'M': 0.031,'I': 12, 'R': 1; 'A': FGSM-LinfAttack,\\'L':DLR$\_$P, 'M': 0.031, 'I': 50, 'R': 0; 'A': CW-LinfAttack, 'M': 0.031,'I': 12, 'R': 1}     \\
			\midrule
			%			Pre-training & WRN-34-10\\
			%			{Proxy}& ResNet-18\\
			%			AT$\_$HE &WRN-34-20\\
			%			LBGAT &WRN-34-20\\
			%			FAT &WRN-34-10\\
			%			Overfitting &WRN-34-20\\
			%			Self-adaptive &WRN-34-10\\
			TRADES \cite{zhang2019theoretically} & WRN-34-10 &  \makecell[c]{'A': MT-LinfAttack,'L':Hinge, 'M': 0.5,'I': 100, 'R': 0; 'A': PGD-LinfAttack,\\'L':CE, 'M': 0.5, 'I': 25, 'R': 1; 'A': DDNL2Attack, 'L':DLR$\_$P,'M': 0.5,'I': 125, 'R': 1}     \\
			\midrule
			%			LBGAT & WRN-34-10\\
			%			OAAT & ResNet-18\\
			%			SAT & WRN-34-10\\
			%			Robustness &ResNet-50\\
			%			YOPO & WRN-34-10\\
			%			MMA & WRN-28-4\\
			%			DNR & ResNet-18\\
			%			CNL & ResNet-18\\
			Feature Scatter \cite{zhang2019defense} & WRN-28-10  & \makecell[c]{'A': MT-L2Attack,'L':CE, 'M': 0.5,'I': 100, 'R': 0; 'A': PGD-L2Attack,\\'L':CE, 'M': 0.5, 'I': 25, 'R': 0; 'A': DDNL2Attack,'DLR$\_$P', 'M': 0.5,'I': 12, 'R': 1}     \\
			\midrule
			Interpolation \cite{zhang2019adversarial} & WRN-28-10  & \makecell[c]{'A': MT-L2Attack, 'L':CE, 'M': 0.5,'I': 100, 'R': 0; 'A': PGD-L2Attack,\\'L':CE, 'M': 0.5, 'I': 25, 'R': 0; 'A': DDNL2Attack, 'L':DLR$\_$P, 'M': 0.438,'I': 125,\\ 'R': 0;'A': MT-L2Attack, 'L':CE, 'M': 0.5,'I': 175, 'R': 3}     \\
			\midrule
			%			Sensible &WRN-34-10\\
			Regularization \cite{jin2020manifold} & ResNet-18  & \makecell[c]{'A': MT-L2Attack, 'L':CE, 'M': 0.5,'I': 100, 'R': 0; 'A': PGD-L2Attack,\\'L':DLR, 'M': 0.375, 'I': 25, 'R': 1; 'A': DDNL2Attack,'L':DLR$\_$P, 'M': 0.5,'I': 125, \\'R': 2;'A': MT-L2Attack, 'L':L1$\_$P, 'M': 0.5,'I': 175, 'R': 0}     \\

			\bottomrule		
	\end{tabular}}
\end{table*}

\subsection{The performance of AAA compared with exsiting manually designed methods}
\label{43}
To illustrate the effectiveness of AAA, we first make the comparison between AAA and existing manually-designed adversarial attacks. The threat models are defensed by various techniques, including PGDAT \cite{pgd2018}, AWP \cite{wu2020adversarial}, MART \cite{wang2019improving}, TRADES \cite{zhang2019theoretically}, Feature Scatter \cite{zhang2019defense}, Interpolation \cite{zhang2019adversarial}, Regularization \cite{jin2020manifold}, Fixing data \cite{rebuffi2021fixing}, OAAT \cite{addepalli2021towards}, LBGAT \cite{cui2021learnable}, Overfit \cite{rice2020overfitting}, IAR \cite{bernhard2020luring} and fast adversarial training (FastAT) \cite{wong2020fast}. The searched adversarial attack according to each defensed model is denoted as AAA. The attack searched on one model and tanfered to other models is denoted as AAA$_{sub}$.

\begin{figure*}[h]
	\centering
%	\subfigure[Interpolation]{
%		\includegraphics[height=0.3\linewidth]{figure/adv_inter.jpg}
%	}
	\subfigure[MART]{
		\includegraphics[height=0.3\linewidth]{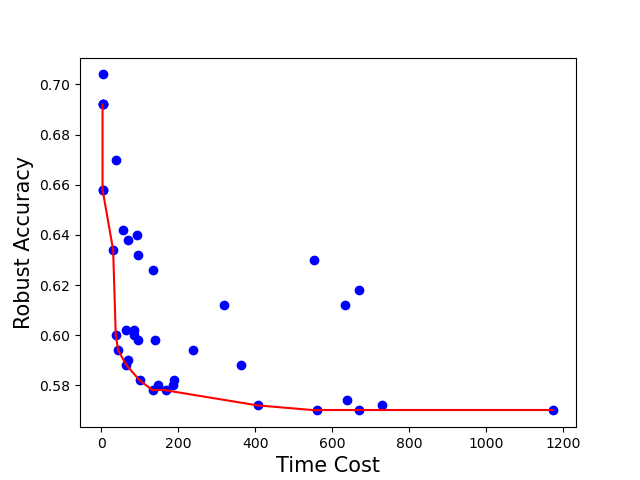}
	}
	\subfigure[TRADES]{
		\includegraphics[height=0.3\linewidth]{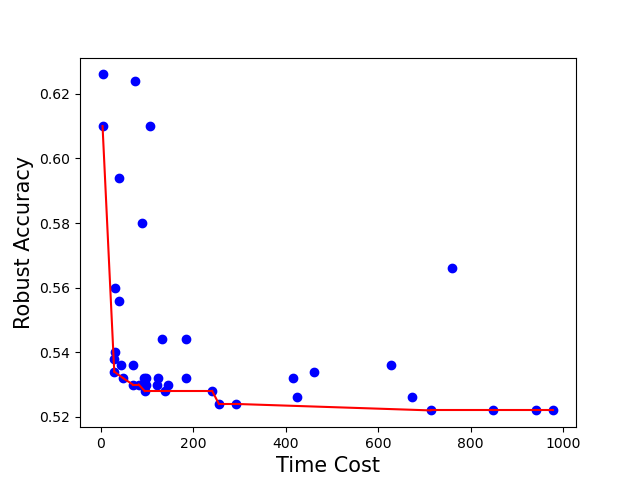}
	}

	\subfigure[PGDAT]{
		\includegraphics[height=0.3\linewidth]{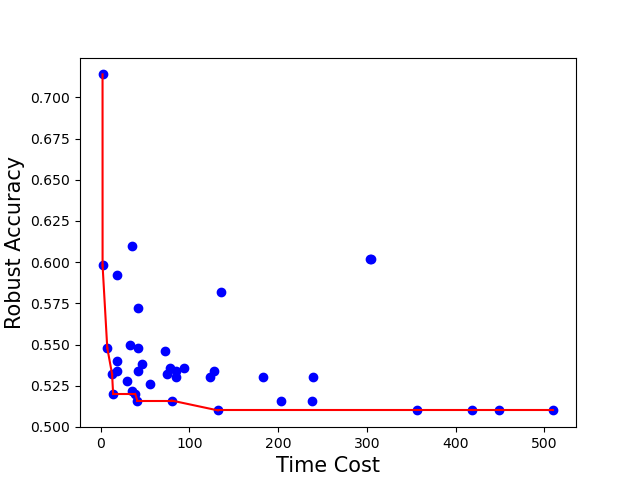}
	}
	\subfigure[Feature Scatter]{
	\includegraphics[height=0.3\linewidth]{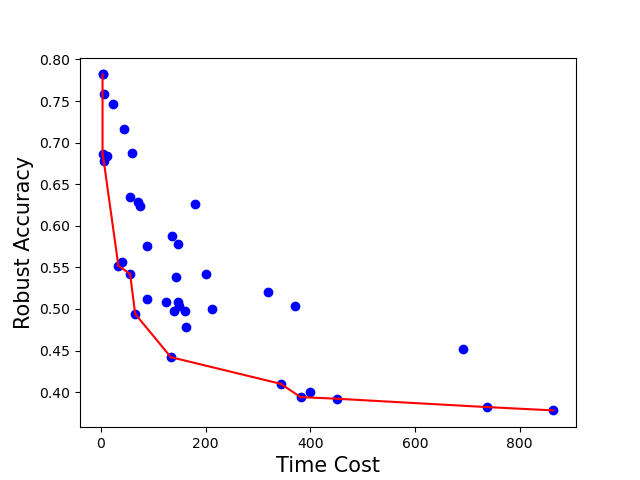}
}	
%	\subfigure[Regularization]{
%		\includegraphics[height=0.3\linewidth]{figure/adv_regular.jpg}
%	}
%	\subfigure[Feature$\_$Scatter]{
%		\includegraphics[height=0.3\linewidth]{figure/Feature_Scatter.png}
%	}
	\caption{The pareto front of the searched attack on various defense methods.}
	\label{nsga}
\end{figure*}

The robust accuracy and time cost under the $l_{\infty}$ attack of APGD, AA and AAA across various defense strategies on CIFAR10, CIFAR100 and ImageNet dataset are presented in Table \ref{cifar10linf}, Table \ref{cifar10l2}, Table \ref{cifar100linf} and Table \ref{ImageNetlinf} respectively. The searched pareto front of the adversarial attack on each defensed model is visualized in Fig \ref{nsga}. The final searched attacks on the CIFAR10 dataset are visualized in Table \ref{linf} and Table \ref{l2}. In our visualization for the loss function, we use the original representation if the probability output is adopted. For example, DLR stands for the loss function of $l_{\text{DLR}}$ with probability output and DLR$\_$P represents that of the $l_{\text{DLR}}$ with the logit output.

From Fig \ref{nsga}, it can be seen that the design of auto adversarial attack is a typical multi-objective optimization problem, and our proposed algorithm could obtain a competitive pareto front on various defensed models. The adversarial attack with the lowest robust accuracy and the least time cost can be regarded as the near optimum of our proposed optimization problem. 

\begin{figure*}[h]
	\centering
	\subfigure[CAA]{
		\includegraphics[height=0.3\linewidth]{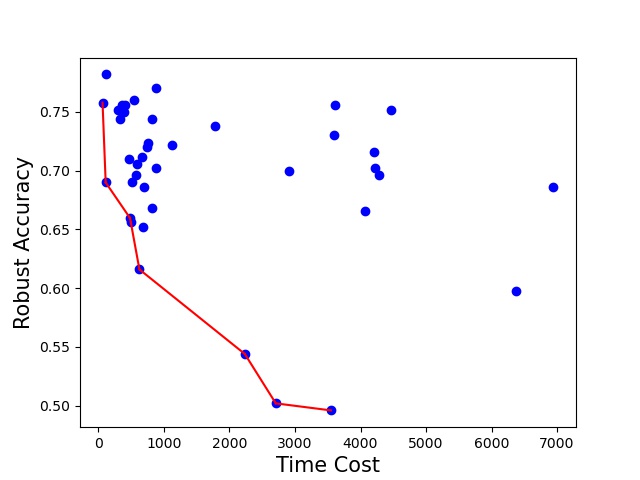}
	}
	\subfigure[AAA]{
		\includegraphics[height=0.3\linewidth]{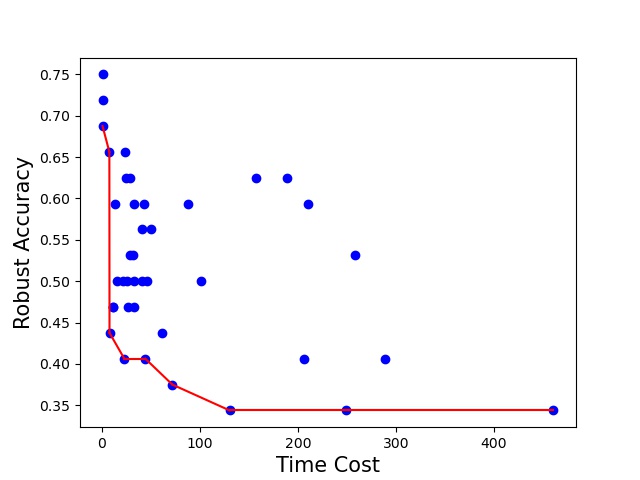}
	}
	\caption{The comparison of the distribution of the population initialization by AAA and CAA.}
	\label{AAAinitial}
\end{figure*}

From Table \ref{cifar10linf}, we can see that both APGD and AA achieve the competitive performance with robust accuracy. AA obtains lower robust accuracies on all the defensed models than APGD, which proves the effectiveness of ensembling the advantages of different adversarial attacks. But the evaluation of AA also needs to take a longer time, which takes over 100 minutes on one defensed model on average. AAA$_{sub}$ is searched on the PGDAT model and evaluated on other defensed models. It can be seen that with the similar robust accuracy, the time cost of AAA$_{sub}$ is largely decreased. For example, as for PGDAT and MART models, AAA$_{sub}$ can reduce the time cost of evaluation from 90.98min and 123.96min in AA to 10.13min and 19.10min, respectively. In addition, on the regularization model, the robust accuracy of AAA$_{sub}$ is only 3.6$\%$, farther lower than AA with 6$\%$. It shows that the searched ensemble attack can be more efficient than the fixed one in AA. Besides, the searched attack possesses superior transferability. Thus the attack can be searched on one typical defense method and utilized to generate adversarial examples on other models.

Table \ref{cifar10l2}, Table \ref{cifar100linf} and Table \ref{ImageNetlinf} show the results of APGD, AA, AAA$_{sub}$ and AAA under $l_{2}$ attack on CIFAR10 and $l_{\infty}$ attack on other datasets. On $l_{2}$ attack on CIFAR10, AAA$_{sub}$ achieves the better robust accuracy performance on three defensed models than AA. By searching the attack on each defensed models, the performance can be further improved, where AAA achieves the best robust accuracy performance on seven of eight defensed models. On $l_{\infty}$ attack on CIFAR100 and ImageNet datasets, AAA also outperforms AA and AAA$_{sub}$ on multiple defensed models. From Table \ref{linf} and Table \ref{l2}, we can see that the superior attack such as MT-Attack plays an key role in the final performance of the searched attack, and some weaker attacks such as FGSM-Attack can also improve the attack performance. These above results show the competitive performance of our proposed AAA compared with existing manually-designed attacks.

\subsection{The performance of AAA compared with CAA}

In this section, the effectiveness of AAA compared with CAA is illustrated. To provide the fair comparison environment, we implement them on the same computational device. As reported in their paper, the multi-objective optimization algorithm without local search is adopted to solve CAA. We make the comparison from three parts including the population initialization, the search cost and the performance of the searched attack.
\begin{table*}[h]
	\renewcommand\arraystretch{1.5}
	\scriptsize
	\centering
	\caption{The robust accuracy ($\%$) of the searched attack of AAA, CAA$_{}^{*}$ and CAA$\_{sub}$.}
	\label{search_time}
	\setlength{\tabcolsep}{4mm}{
		\begin{tabular}{c|c|ccccc}
			
			\toprule
			
			\multirow{1}{*}{Dataset}&	\multirow{1}{*}{Defense Method}&	\multirow{1}{*}{Model}&\multicolumn{1}{c}{CAA$_{}^{*}$} &\multicolumn{1}{c}{CAA$\_{sub}$}& \multicolumn{1}{c}{AAA}& \\
			
			\midrule			
			CIFAR10($l_{\infty}$)	&	{PGDAT \cite{pgd2018}}&	{ResNet-50} & 51.2 & 51 &  \textbf{50.6}    \\	
			%			{ULAT \cite{gowal2020uncovering}}&	{WRN-70-16}   &                     \\	
			%			{Fixing Data \cite{rebuffi2021fixing}}	&	{WRN-70-16}            \\	
			%			{ULAT \cite{gowal2020uncovering}}	&	{WRN-28-10}                   \\	
			%		 	{Fixing Data \cite{rebuffi2021fixing}}	&	{WRN-28-10}                     \\
			%			{RLPE \cite{sridhar2021robust}}&	\text { WRN-34-15 }   \\
			%		{Geomoetry}	&	\text {WRN-28-10 }        \\
			&	{AWP \cite{wu2020adversarial}}	&	\text {WRN-34-10 }  & 55.2  &55 &\textbf{54.8} \\
			%			{RST}	&	\text { WRN-28-10 }                       \\
			
			%			{Proxy}&	{WRN-34-10}   &         \\	
			%				{OAAT} &    WRN-34-10  &            \\	
			%			HYDRA &	{WRN-28-10}                                                        \\	
			%			ULAT&	{WRN-70-16}                  \\
			& MART \cite{zhang2019theoretically} & WRN-28-10 &\textbf{57.0}&57.2 & \textbf{57.0} \\
			
			&	Feature Scatter \cite{zhang2019defense}  & WRN-28-10 &37.2& 37.6 &\textbf{36.6}  \\
			&	Interpolation \cite{zhang2019adversarial} & WRN-28-10 &44.0&36.4 &\textbf{35.8} \\
			%			Sensible &WRN-34-10\\
			&	Regularization \cite{jin2020manifold} & ResNet-18 &3.0& 3.8 & \textbf{1.88}\\	
			
			\midrule			
			CIFAR10($l_{2}$)	&	{PGDAT \cite{pgd2018}}&	{ResNet-50} & 58.8 & 58.0 &  \textbf{57.2}    \\	
			%			{ULAT \cite{gowal2020uncovering}}&	{WRN-70-16}   &                     \\	
			%			{Fixing Data \cite{rebuffi2021fixing}}	&	{WRN-70-16}            \\	
			%			{ULAT \cite{gowal2020uncovering}}	&	{WRN-28-10}                   \\	
			%		 	{Fixing Data \cite{rebuffi2021fixing}}	&	{WRN-28-10}                     \\
			%			{RLPE \cite{sridhar2021robust}}&	\text { WRN-34-15 }   \\
			%		{Geomoetry}	&	\text {WRN-28-10 }        \\
			&	{AWP \cite{wu2020adversarial}}	&	\text {WRN-34-10 } &  62.4 &60.0  &\textbf{59.6}\\
			%			{RST}	&	\text { WRN-28-10 }                       \\
			
			%			{Proxy}&	{WRN-34-10}   &         \\	
			%				{OAAT} &    WRN-34-10  &            \\	
			%			HYDRA &	{WRN-28-10}                                                        \\	
			%			ULAT&	{WRN-70-16}                  \\
			& MART \cite{zhang2019theoretically} & WRN-28-10& 63.0&62.8 & \textbf{62.4} \\
			&	TRADES \cite{zhang2019theoretically} & WRN-34-10&59.2 & 58.8 & \textbf{58.2} \\
			&	Feature Scatter \cite{zhang2019defense}  & WRN-28-10&53.6 & 53.8 & \textbf{52.0} \\
			&	Interpolation \cite{zhang2019adversarial} & WRN-28-10&46.6 &45.8 & \textbf{42.8}\\
			%			Sensible &WRN-34-10\\
			&	Regularization \cite{jin2020manifold} & ResNet-18&26.8 & 27.4 & \textbf{25.8}\\					
			\midrule			
			CIFAR100($l_{\infty}$)	&		{Fixing Data \cite{rebuffi2021fixing}}	&	{WRN-28-10}  & 31.8 &31.6 &   \textbf{31.4}   \\	
			
			&{OAAT \cite{addepalli2021towards}}	&	{WRN-34}   & 33.2  &32.8 & \textbf{32.6}\\
			&{OAAT \cite{addepalli2021towards}}	&	{ResNet-18}  &  27.6  &27.0 &\textbf{26.6} \\
			& {LBGAT \cite{cui2021learnable}}	&	\text {WRN-34-10 } &30.6 &\textbf{30.4}& \textbf{30.4} \\
			
			&	{AWP \cite{wu2020adversarial}}	&	\text {WRN-34-10 }&30.2 & \textbf{30.0} & \textbf{30.0}\\
			&	{IAR \cite{bernhard2020luring}}	&	\text {WRN-34-10 } &25.0 &25.4  &\textbf{24.8}\\
			%			&	{ULAT \cite{gowal2020uncovering}}&	{WRN-70-16} &30.2 &\\
			&	{Overfit \cite{rice2020overfitting}}	&	\text {PAResNet-18 } &20.8& 20.6 &\textbf{20.4} \\				
			\bottomrule		
	\end{tabular}}
\end{table*}
Population initialization is the key step in multi-objective optimization. In CAA, given the maimum attack magnitude, the search space is uniformly divided into 8 parts. Different configuration of magnitudes is obtained in the population initialization. In our proposed AAA, all of the magnitudes are set as the maximum values in the primary initialization, which could alleviate the problem that the magnitudes of initial individuals are farther lower than the predefined maximum value. As shown in Fig. \ref{AAAinitial}, we present the robust accuracy and time cost of initial individuals after the population initialization. It could be seen that AAA could obtain more individuals with lower robustness accuracy and less time cost.
%\right) 
The comparison of the search cost between AAA and CAA on one defensed model is listed in Table~\ref{ra}. It could be seen that AAA is ten times faster than CAA due to the strategy of constructing the rough evaluator.

\begin{table}[h]
	\renewcommand\arraystretch{1.5}
	\scriptsize
	\centering
	\caption{The comparison of the searching time of AAA and CAA.}
	
	\label{ra}
	\setlength{\tabcolsep}{4mm}{
		\begin{tabular}{ccc}
			
			\toprule
			Method & \multicolumn{1}{c}{CAA} & \multicolumn{1}{c}{AAA}\\
			%			\multirow{2}{*}{CIFAR10}&	\multirow{2}{*}{Model} & &\multicolumn{2}{c}{AAA} \\
			%			
			%			\cmidrule(r){3-3}  \cmidrule(r){4-4}  \cmidrule(r){5-6}  \cmidrule(r){7-8}
			%			\cmidrule(r){9-10}
			%			&	& acc & acc	 &   acc		
			%			&  time cost     &  acc   &   time cost &  acc   &   time cost \\		
			\midrule		
			The searching time cost & 20GPU days & \textbf{3GPU days}\\	
			\bottomrule		
	\end{tabular}}
\end{table}

Lastly, we compare the performance of the searched adversarial attack by AAA and CAA. Due to the reason that the authors did not provide the source code of multi-objective search, we compare the searched attack reported by them, the searched attack of CAA implemented by us, the searched attack of AAA. These three attacks are denoted as CAA$_{sub}$, CAA$^{*}$, and AAA respectively. To provide a fair comparison environment, in CAA$^{*}$, the length of attack sequence is set to four, and the parameter setting in NSGA-II is the same as AAA. Their attack performances on various defense methods are listed in Table~\ref{search_time}.

From Table~\ref{search_time}, it can be seen that AAA outperforms two other adversarial attacks on all of three tasks, including $l_{2}$, $l_{\infty}$ attack on CRIFA10 and $l_{\infty}$ attack on CRIFA100 dataset. On all the defensed models, AAA achieves the loweset robust accuracy, while CAA$^{*}$ only obtain the lowest robust accuracy on one model and CAA$_{sub}$ obtain the lowest robust accuracy on two models, respectively. In addition, on some defensed models, the improvement of AAA compared with CAA$^{*}$ and CAA$_{sub}$ is relatively large. For example, on the model defensed by regularization, CAA$^{*}$ and CAA$_{sub}$ only obtain the robust accuracy of 3.0$\%$ and 3.8$\%$ on $l_{\infty}$ attack of CIFAR10 dataset respectively, while our AAA can reduce it to 1.88$\%$. Similar phenomennon can also be seen on $l_{2}$ attack.

\subsection{The performance of the search space} 

\begin{table*}
	\renewcommand\arraystretch{1.5}
	\scriptsize
	\centering	
	\caption{The comparison of different restarts in an adversarial attack with length of three on CIFAR100.}
	\setlength{\tabcolsep}{4mm}{
		\begin{tabular}{ccc|ccc}
			
			\toprule
			Restart & \multicolumn{1}{c}{RA($\%$)} & \multicolumn{1}{c}{Time(s)}&Restart & \multicolumn{1}{c}{RA($\%$)} & \multicolumn{1}{c}{Time(s)}\\
			%			&   time cost &  acc   &   time cost \\		
			\midrule		
			$[0,0,0]$ & 27.0 & 302.99&$[0,0,2]$ &27.0  & 287.64\\
			%			\midrule
			$[0,1,2]$ &27.0  & 258.39&$[0,1,0]$ & \textbf{26.8} & 273.07\\
			%			\midrule		
			$[0,0,1]$ &27.0  & 287.92&$[0,1,1]$ & 27.0 & \textbf{257.06}\\	
			\bottomrule		
	\end{tabular}}
	\label{restart}
\end{table*}

\begin{table*}[t]
	%	\renewcommand\arraystretch{1.5}
	%	\scriptsize
	\centering
	\caption{The comparison of different restarts in an adversarial attack with length of four on CIFAR10.}
	\label{restart4}
	\setlength{\tabcolsep}{4mm}{
		\begin{tabular}{ccc|ccc}
			
			\toprule
			Restart & \multicolumn{1}{c}{RA($\%$)} & \multicolumn{1}{c}{Time(s)}&Restart & \multicolumn{1}{c}{RA($\%$)} & \multicolumn{1}{c}{Time(s)} \\		
			\midrule		
			$[0,0,0,0]$ & 3.6 & 711.13&$[0,1,0,0]$ & 4.0& 502.07\\
			%			\midrule
			$[0,0,0,1]$ &3.6  & 609.24 &$[0,1,0,1]$ & 3.8 & 451.01\\
			%			\midrule		
			$[0,0,0,2]$ & 3.6 & 812.46 &$[0,1,0,2]$ & 4.2 &450.26\\
			%			\midrule
			$[0,0,0,3]$ & 3.8  & 531.39&$[0,1,0,3]$ & 3.8 & 506.28\\
			%			\midrule		
			$[0,0,1,0]$ &3.6 & 505.66&$[0,1,1,0]$ & 4.0 &456.50\\
			%			\midrule
			$[0,0,1,1]$ & 3.6 &443.48&$[0,1,1,1]$ & \textbf{3.0} & 384.82\\
			%			\midrule		
			$[0,0,1,2]$ & 3.4  & 448.80 &$[0,1,1,2]$ & 3.2 & 378.00\\
			%			\midrule
			$[0,0,1,3]$ &3.8  & 388.67&$[0,1,1,3]$ & 4.6 & 398.46\\
			%			\midrule
			$[0,0,2,0]$ & \textbf{3.0} & 512.97&$[0,1,2,0]$ & 5.0 &444.02\\
			%			\midrule
			$[0,0,2,1]$ & 3.8 & 448.88&$[0,1,2,1]$ & 3.6 & 375.66\\
			%			\midrule		
			$[0,0,2,2]$ & 4.6 & 507.78 &$[0,1,2,2]$ & 4.6 & 388.78\\
			%			\midrule
			$[0,0,2,3]$ &3.4 & 505.33&$[0,1,2,3]$ & 3.6 & \textbf{371.87}\\	
			\bottomrule		
	\end{tabular}}
\end{table*}

To illustrate the effectiveness of our proposed search space, an adversarial attack sequence with the length of three on the CIFAR100 dataset is taken as an example. The randomly selected adversarial attack is visualized as [{'A': 'MT-LinfAttack', 'L': 'CE', 'M': 0.031, 'I': 50.0, 'R': 0}, {'A': 'MT-LinfAttack', 'L': 'CE', 'M': 0.031, 'I': 25.0, 'R': 0}, {'A': 'CW-LinfAttack', 'L': 'DLR$\_$P', 'M': 0.031, 'I': 125.0, 'R': 0}]. The restart of the adversarial attack sequence is changed to all possible configurations. There are three attack cells in this attack sequence, which possesses six configurations of the restart. The robust accuracy and time cost of the adversarial attack with different restarts are evaluated on the Resnet-18 model defensed by OAAT. The compasrion of different restart is listed in Table~\ref{restart}.

To further illustrate the influence of the restart, we take an adversarial attack with the length of four on the CIFAR10 dataset as another example. The adversarial attack is visualized as [{'A': 'MT-LinfAttack', 'L': 'CE', 'M': 0.031, 'I': 50, 'R': 0}, {'A': 'PGD-LinfAttack', 'L': 'CE', 'M': 0.031, 'I': 25, 'R': 0}, {'A': 'CW-LinfAttack', 'L': 'L1', 'M': 0.031, 'I': 125, 'R': 0}, {'A': 'MI-LinfAttack', 'L': 'DLR', 'M': 0.031, 'I': 125, 'R': 0}]. Compared with the above example, there are more possible configurations of the restart. Similarly, the robust accuracy and time cost of all possible adversarial attacks with different restarts are evaluated on the resnet18 model defensed by regularization. The compasrion of them is presented in Table~\ref{restart4}.

In Table~\ref{restart} and Table~\ref{restart4}, [0,0,0] and [0,0,0,0] represent the connection way of AA, while [0,1,2] and [0,1,2,3] stands for that of CAA. From Table~\ref{restart}, it could be seen that both [0,0,0] and [0,1,2] achieve the robust accuracy of 27.0$\%$, while [0,1,0] could obtain lower robust accuracy of 26.8$\%$. In addition, the time cost of [0,1,2] is 258.39s, which is less than [0,0,0] with 302.99s. It proves that generating the adversarial example based on the previous perturbed image could help to enhance the efficiency. Compared with [0,0,0] and [0,1,2], the better connection way of different adversarial cell is [0,1,1], which needs 257.06s. The gap in the time cost between different restarts increase with the number of test examples. A similar phenomenon can also be seen in Table~\ref{restart4}. It could be seen that both [0,0,0,0] and [0,1,2,3] achieve the robust accuracy of 3.6$\%$, while both [0,1,1,1] and [0,0,2,0] could obtain lower robust accuracy of 3.0$\%$. In addition, the time cost of [0,1,1,1] and [0,0,2,0] are only 384.46s and 512.97s, which is farther less than [0,0,0,0] with 711.13s. It means that by searching for the best restart, we can find the better connection ways of different adversarial attack cells to obtain lower robust accuracy with less time cost.

\begin{figure}[h]
	\centering
	\includegraphics[scale=0.52]{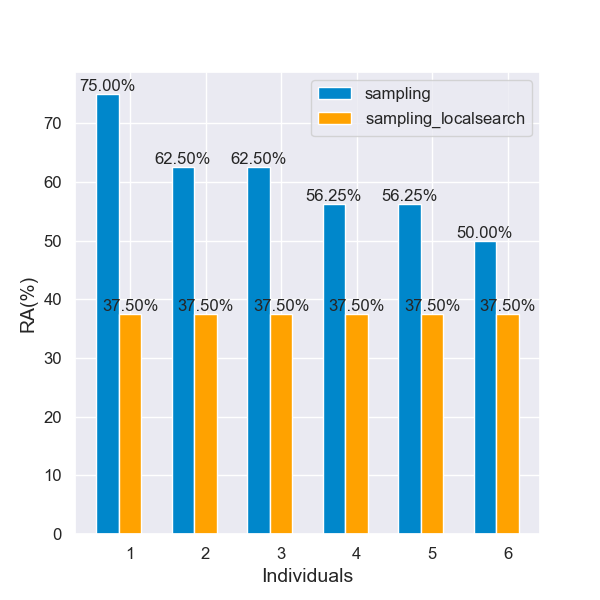}
	\caption{The performance of local search in NSGA-II.}
	\label{local_search}
\end{figure}

\begin{figure*}[htp]
	\centering
	\subfigure[The rank ability using random selection]{
		\includegraphics[height=0.32\linewidth]{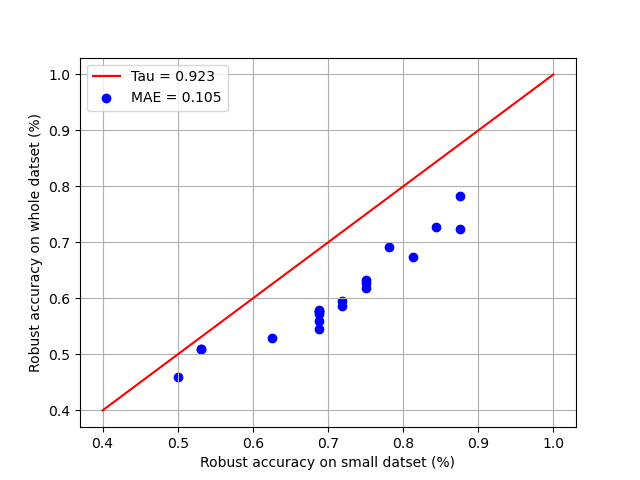}
	}
	\subfigure[The rank ability using loss values]{
		\includegraphics[height=0.32\linewidth]{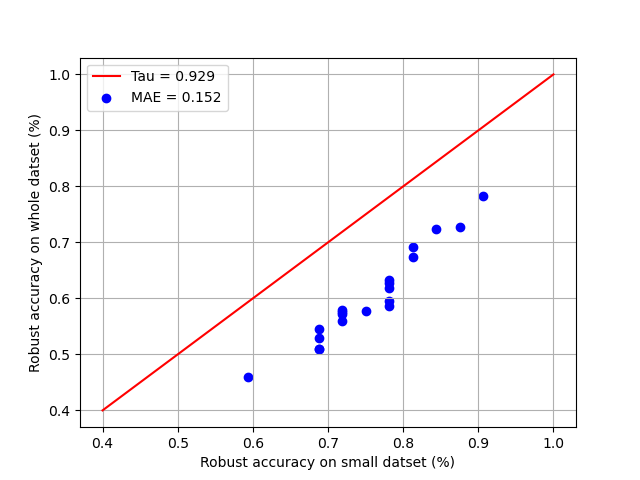}
	}
	\caption{The comparison of the Kendall tau coefficients of robust accuracy by loss values and random order.}
	\label{ra_tau}
\end{figure*}

\begin{figure*}[htp]
	\centering
	\subfigure[The rank ability using random selection]{
		\includegraphics[height=0.32\linewidth]{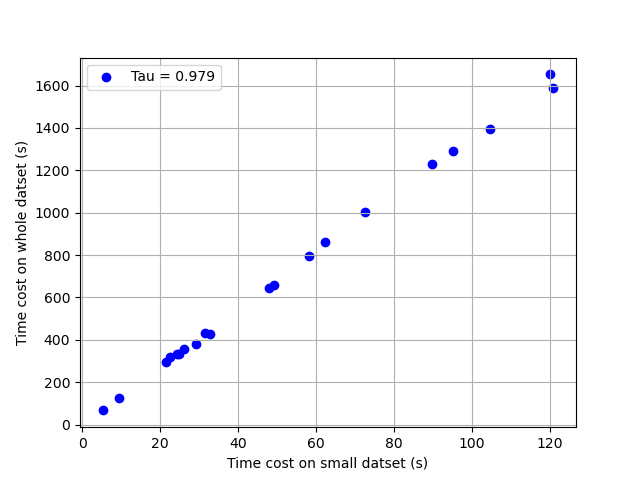}
	}
	\subfigure[The rank ability using loss values]{
		\includegraphics[height=0.32\linewidth]{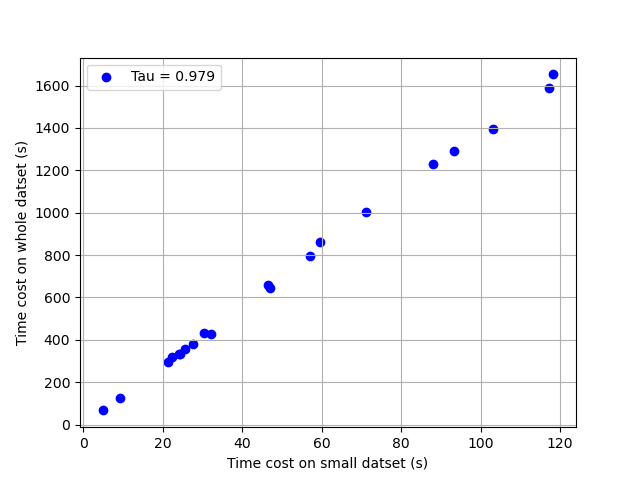}
	}
	\caption{The comparison of the Kendall tau coefficients of time cost by loss values and random order.}
	\label{time_tau}
\end{figure*}

\subsection{The performance of the local search}

 In this section, the effectiveness of the local search to improve the optimization performance is shown.

At the early optimization stage by NSGA-II, we sample some individuals and study the performance of local search, each one of which represents an adversarial attack. In our experiment, six individuals in the first pareto are selected, and their robust accuracies are listed in Fig. \ref{local_search}. The local search is conducted on these six individuals, respectively. The iteration number of local search $T_{l}$ is set to one. As shown in Fig. \ref{local_search}, the robust accuracies of them after local search all decrease to 37.5$\%$. Thus the local search within a relatively small search space could accelerate the process of NSGA-II optimization and help to find better solutions. Particularly for our proposed the connection way of different attack cells, it is more efficient for finding the optimal restart to use local search other than evolutionary based algorithms.

\subsection{The performance of the representative data}

To verify the effectiveness of selecting the representative data using loss values, we adopt the Kendall tau coefficient to evaluate the similarities of the performance using the accurate evaluator and rough evaluator. We randomly generate 20 adversarial attacks and evaluate the robust accuracy and time cost of them on the rough evaluator and accurate evaluator. The comparisons of the Kendall tau coefficients using loss values and random order are presented in Fig.~\ref{ra_tau} and Fig.~\ref{time_tau}.

From Fig.~\ref{ra_tau}, it could be seen that both of the tau coefficients are over 0.92, which verifies the effectiveness of the rough evaluator. The mean absolute error (MAE) between the robust accuracy calculated by the rough evaluator using random order and loss values is about 0.1-0.2$\%$. The MAE of the random order is fewer higher than the accurate evaluator. It is because the points calculated by the rough evaluator with loss values are distributed more uniformly than the evaluator with random order. Besides, the tau coefficient of the rough evaluator sorted by loss values is higher than that sorted by the random order. It means that compared with the random order, the rough evaluator sorted by loss values would give a more accurate performance ranking order. Besides, from Fig.~\ref{time_tau}, it can be seen that the predictions of time cost are also accurate, the tau values of which are about 0.98. Thus, at the primary stage of optimization, the rough evaluator could help to find the near-optimal adversarial attack more efficiently.

\section{Conclusion}
\label{sec5}
In this paper, we propose a multi-objective memetic algorithm for auto adversarial attack optimization design. Different from the previous work about the constructed search space in the ensemble of multiple adversarial attacks, we propose a more diverse and efficient mathematical model, which includes not only the attack operation, attack loss functions, attack magnitude, and iteration number of each attack cell but also the connection way of different cells. In addition, to improve the optimization performance, we develop a multi-objective memetic algorithm that combines NSGA-II and local search, which can obtain better solutions than pure NSGA-II or local search. Lastly, we propose a representative data selection strategy, which selects a small number of data from the whole dataset to evaluate the candidate attacks, accelerating the search process greatly. Compared with existing advanced manually-designed attacks such as AA, our proposed method can find more efficient adversarial attacks. In addition, compared with existing search-based methods such as CAA, our search method can reduce the search time from 20GPU days to 3GPU days and obtain better solutions. Experimental results are verified on multiple defensed methods on CIFAR10, CIFAR100, and ImageNet datasets.

\section*{Acknowledgement}

This work was supported in part by National Natural Science Foundation of China under Grant No.11725211, 52005505 and 62001502.

\section*{Conflict of interest statement}
The authors declare that they have no conflict of interest.

\bibliographystyle{unsrt}
%\bibliography{sample_library} 
\bibliography{mybibfile}   % name your BibTeX data base

\end{document}